\documentclass{article}
\usepackage{iclr2021_workshop,times}

\usepackage{amsmath,amsfonts,bm}

\def\eqref#1{equation~\ref{#1}}

\def\1{\bm{1}}

\DeclareMathAlphabet{\mathsfit}{\encodingdefault}{\sfdefault}{m}{sl}
\SetMathAlphabet{\mathsfit}{bold}{\encodingdefault}{\sfdefault}{bx}{n}

\usepackage{hyperref}
\usepackage{url}
\usepackage{graphicx}
\usepackage[utf8]{inputenc}
\usepackage{amsmath}
\usepackage{amsfonts}
\usepackage{booktabs}
\usepackage[switch]{lineno}  %
\usepackage{xcolor}
\usepackage{subfig}
\usepackage{wrapfig}

\title{Bermuda Triangles: GNNs Fail to Detect \\ Simple Topological Structures}

\author{Arseny Tolmachev, Akira Sakai, Masaru Todoriki \& Koji Maruhashi \\
Fujitsu Research, Fujitsu, Ltd.\\
4-1-1 Kamikodanaka, Nakahara-ku Kawasaki-shi, Kanagawa, Japan \\
\texttt{\{t.arseny,akira.sakai,todoriki.masaru,maruhashi.koji\}@fujitsu.com} }

\renewcommand{\cite}[1]{\citep{#1}} 

\iclrfinalcopy 
\begin{document}

\maketitle

\begin{abstract}
Most graph neural network architectures work by message-passing node vector embeddings over the adjacency matrix, and it is assumed that they capture graph topology by doing that.
We design two synthetic tasks, focusing purely on topological problems -- triangle detection and clique distance -- on which graph neural networks perform surprisingly badly, failing to detect those ``bermuda'' triangles.
Datasets and their generation scripts are publicly available on github.com/FujitsuLaboratories/bermudatriangles and dataset.labs.fujitsu.com.
\end{abstract}

\section{Introduction}

\begin{wraptable}{r}{6cm}
\begin{tabular}{lrr}
Method & Triangles & Clique \\
\toprule
GCN & 50.0 & 50.0 \\
GCN+D & 75.7 & 83.2 \\
GCN+D+ID & 80.4 & 83.4 \\
GIN & 74.1 & 97 \\
GIN+D & 75.0 & 99.4 \\
GIN+D+ID & 70.5 & 100.0 \\
GAT & 50.0 & 50.0 \\
GAT+D & 88.5 & 99.9 \\
GAT+D+ID &  94.1 & 100.0 \\
\midrule
SVM+WL & 67.2 & 73.1 \\
SVM+Graphlets & 99.6 & 60.3 \\
\midrule
FCNN & 55.6 & 54.6 \\
TF & 100.0 & 70.0 \\
TF+AM & 100.0 & 100.0 \\
TF-IS+AM & 86.7 & 100.0 \\
TF-IS+AM4 & 97.5 & 100.0 \\
\end{tabular}
\caption{Test accuracy, \% on proposed synthetic datasets. We expect that a method should achieve 100\% accuracy on both datasets.}
\label{tab:synch_datasets}
\vspace{-0.75cm}
\end{wraptable}

Many tasks need to handle the graph representation of data in areas such as chemistry \cite{Wale}, social networks \cite{fan2019graph}, and transportation \cite{Zhao2019}.
Furthermore, it is not limited to these graph tasks but also includes images \cite{ML-GCN_CVPR_2019} and 3D polygons \cite{Point-GNN} that are possible to convert to graph data formats.
Because of these broad applications, Graph  Deep Learning is an important field in machine learning research.

Graph neural networks (GNNs, \cite{scarselli2008graph}) is a common approach to perform machine learning with graphs.
Most graph neural networks update the graph node vector embeddings using the message passing.
Node vector embeddings are usually initialized with data features and local graph features like node degrees.
Then, for a $(n+1)$-th stacked layer, the new node state is computed from the node vector representation of the previous layer ($n$).
There exist many approaches, and most of them follow this big pattern.
Message passing occurs on the graph adjacency matrix and is completely baked in the algorithm.

Deep learning for graphs uses data sets from a variety of fields~\cite{Jure2014snapnets,hu2020open}.
For example, there are protein molecules, social networks, and web networks.
Several tools to investigate and visualize theses graph data are proposed~\cite{Ryan2015graphtool}.
Usually, properties of graphs are measured using graph invariants, i.e., order, size, connectivity, etc.
Applying GNNs to a task that can be solved by a simple classifier with graph invariants as input features is excessive.
This is because there are usually fast and excellent algorithms for calculating graph invariants.
Therefore, it is a very important question whether the target graph problem is hard enough to be a target for the development of graph deep learning, but it has not been investigated.
Researchers in traditional graph algorithms have utilized synthetic data with various characteristics to investigate these issues.
For example, G-set~\cite{stefanxxxgset} is used in the study of the Max-cut problem. This data set is an artificially generated data set with difficult characteristics for the Max-cut problem.
We employ this direction.
\citet{Dehmamy2019UnderstandGNN} gave theoretical insights in graph topology learning power of GNNs.
However, to our best understanding, there were counterexample topological datasets which were not easily solvable by GNNs.

Based on the structure of the algorithms, it is assumed that deeper representations in GNNs capture adjacency information, including graph topology \cite{Xu2019}.
As a counterexample, we propose two seemingly easy synthetic graph classification tasks, focusing solely on graph topology: triangle detection task and clique distance task.
Modern GNNs perform surprisingly poorly on both tasks, see Table~\ref{tab:synch_datasets}, failing to find these ``bermuda'' triangles in the first proposed task.
In addition to that, we suggest an approach that can solve both synthetic tasks perfectly.


\section{Synthetic Tasks} 
\label{sec:synthetic_tasks}

In this section, we describe the proposed synthetic tasks with a way to generate them.
The tasks are detecting a presence of a triangle (loop of a length 3) in a graph (\textbf{Triangle}) or detecting whether a distance between two cliques is lower or higher than a certain threshold (\textbf{Clique Distance}).
Within the generation process, we employ logistic regression-based filtering which removes spurious features from the data, leaving presumably only the features which should be directly related to solving the task at hand.


\subsection{Undermanned Logistic Regression Filtering} 
\label{sub:undermanned_logistic_filtering}

\begin{wrapfigure}{r}{6cm}
\includegraphics[width=0.35\textwidth]{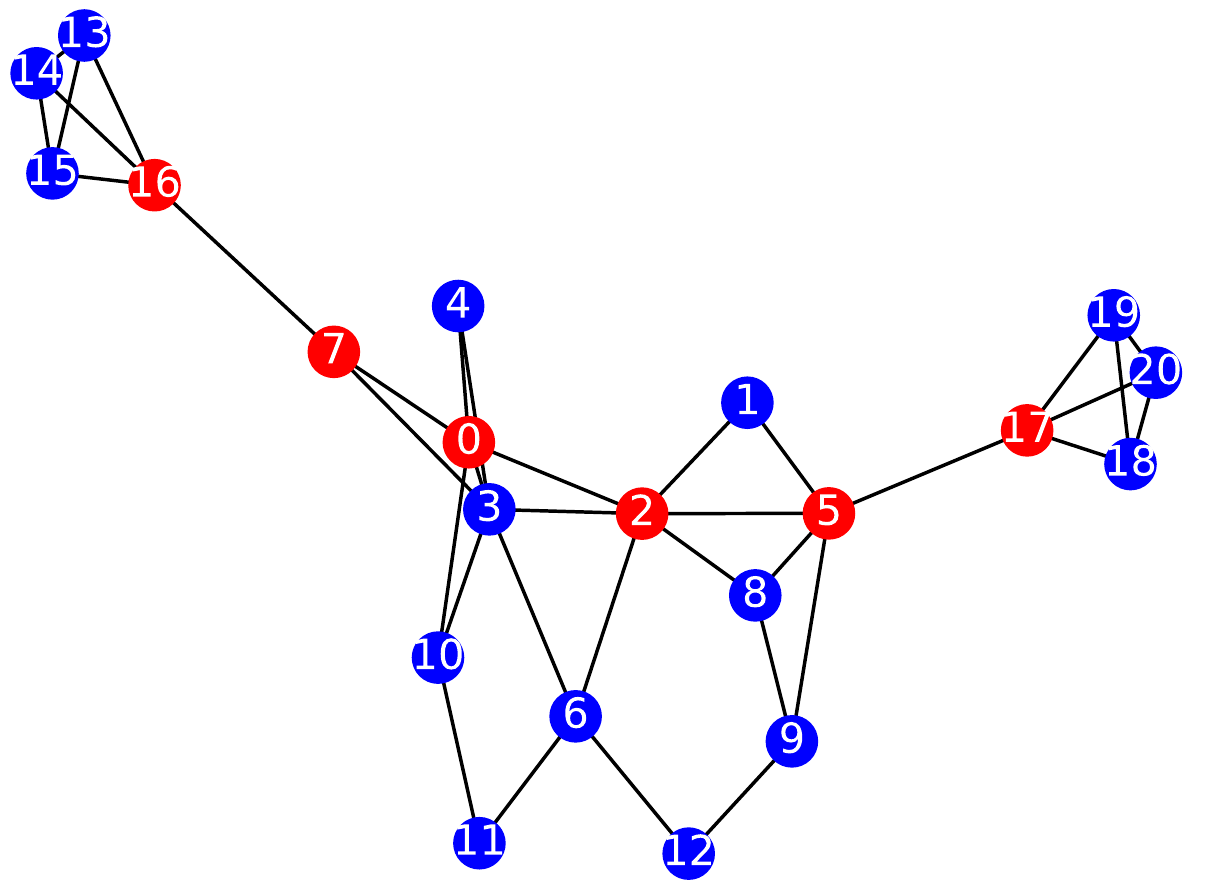}
\caption{An example from Clique distance dataset. The shortest path between cliques is highlighted in red.}
\label{fig:clique_dist}
\vspace{-0.30cm}
\end{wrapfigure}

The objective of synthetic datasets is to check whether a model can process a target phenomenon.
Spurious correlations in synthetically created data interfere with this objective and can be alternatively seen as if the data contains a sub-task that is not one we are interested in.
We employ \emph{undermanned} logistic regression filtering to combat this effect.
We build a logistic regression classifier with a just-not-enough feature template set, so it will be impossible for it to solve the task at hand.
We then select mostly data items that can not be solved with this undermanned classifier.
Perfectly, such a classifier should have low accuracy even on unfiltered data, but that can be not the case for randomly generated data.

The filtering procedure builds on the overlapping $n$-fold cross-validation.
First, similarly to standard cross-validation, we split the dataset into $n$ folds.
We train a classifier on $m < n$ folds and use the remaining $n - m$ for the validation.
By repeating the training-validation process $n$ times, shifting folds each time in a round-robin fashion, we can compute for each data item the number of times the classifier was able to produce the correct result.
Intuitively, we are not interested in data for which the classifier was always correct, so we sample the final dataset by biasing the items which had at least one classification error.

We use the following feature templates for both of our datasets:
1) Node degree (number of edges),
2) Node degrees of both edge ends,
3) Number of nodes in the graph,
4) Number of edges in the graph.
The first two feature templates are used two times: to check both the presence of a feature (e.g., there exists a node with a degree of five) and its count.
This feature set should not be enough to detect the presence of a triangle in the data.
Still, the logistic regression classifier has the accuracy of $82.5\%$ on the unfiltered data for the clique distance problem and $87\%$ for the triangle problem.
We generate 200k graph candidates (100k for each class) and filter them to produce 10k train data and 1k test data.

\begin{figure}
    \centering
    \includegraphics[width=.35\textwidth]{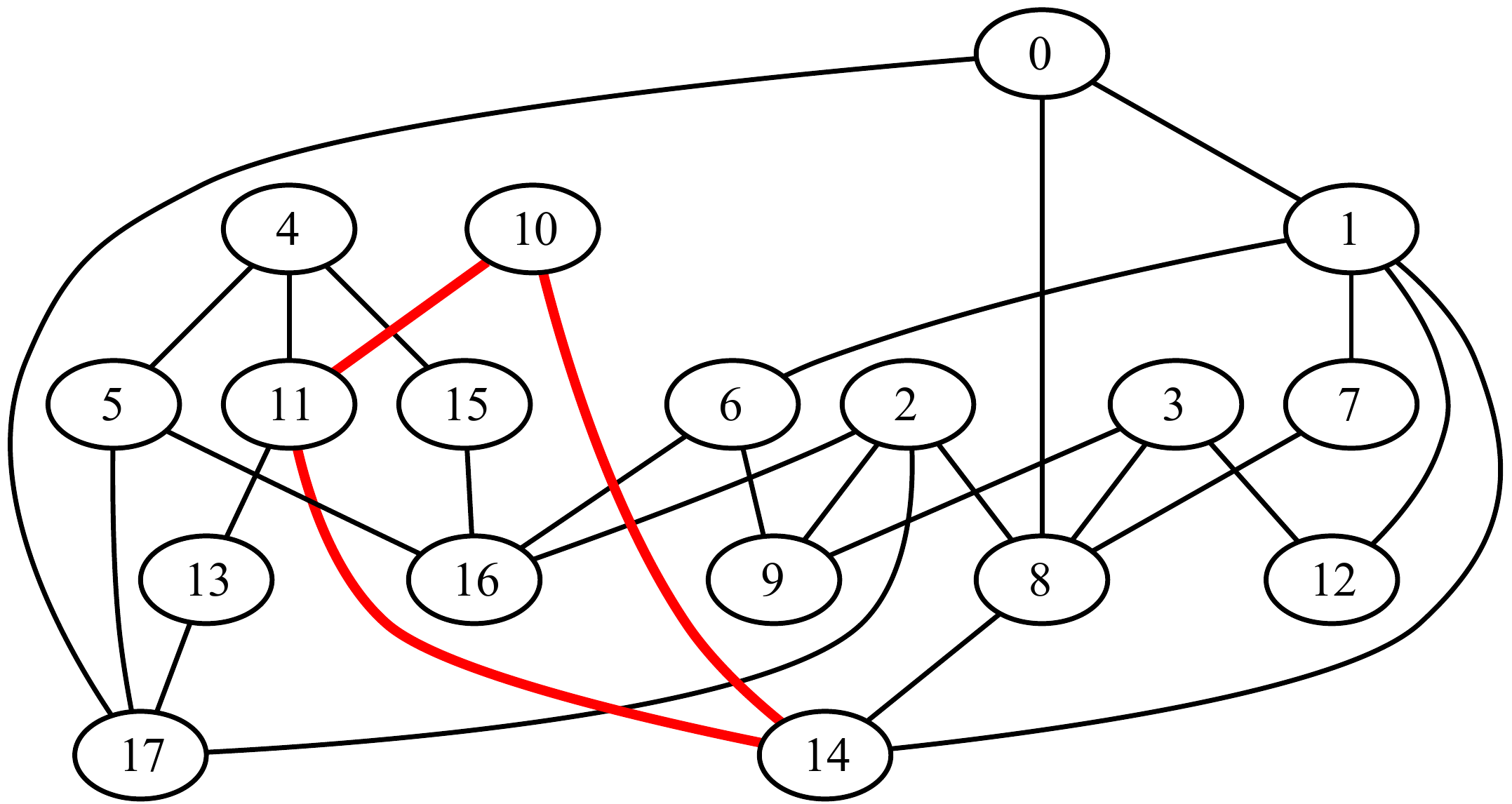}
	\includegraphics[width=.35\textwidth]{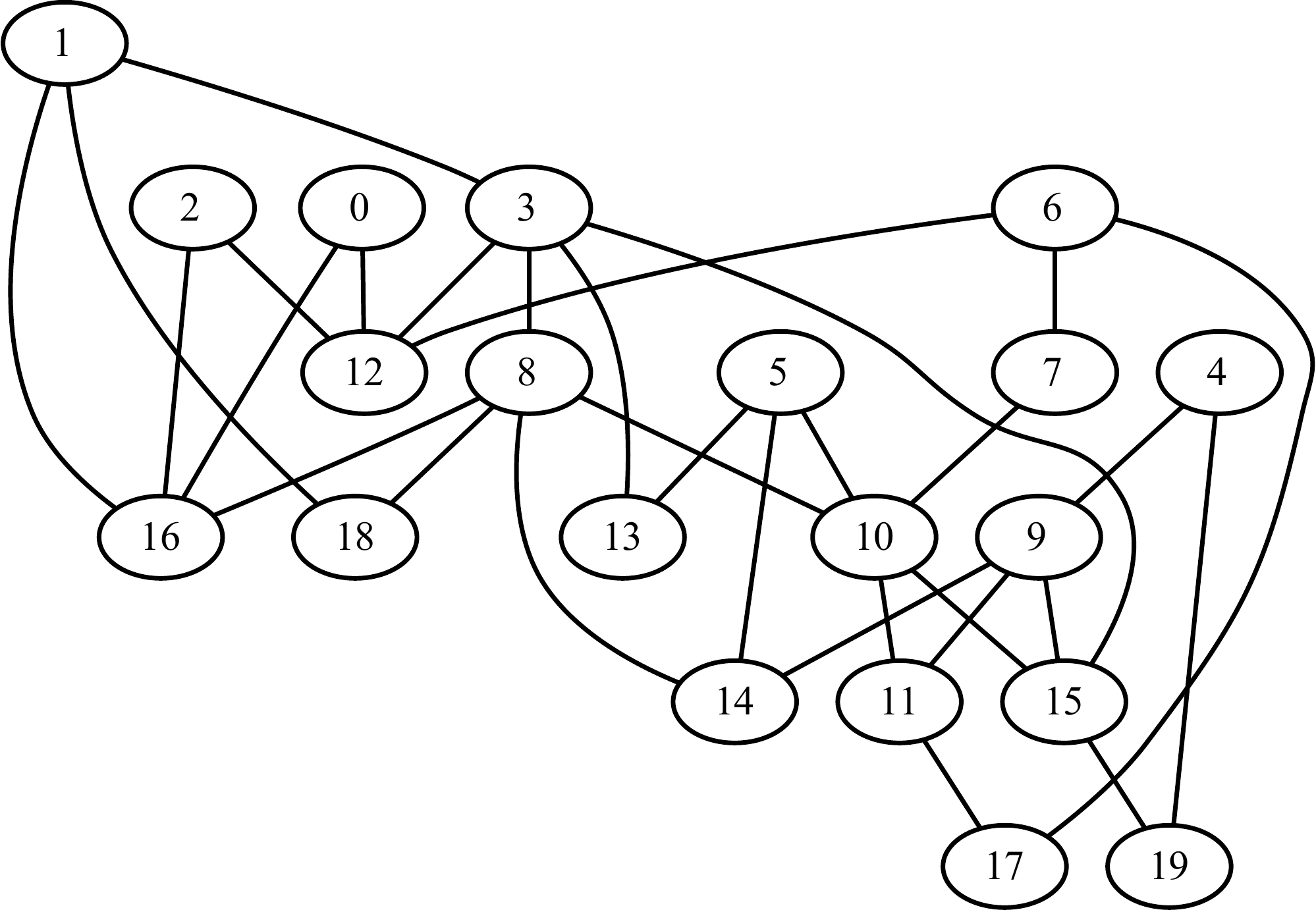}
    \caption{Two data examples from the Triangles dataset.
	The left one has a triangle highlighted in bold red.
	The right one does not have any triangles.}
    \label{fig:triangles}
\end{figure}

\subsection{Triangles Dataset}

In the Triangles dataset, each data item is a graph that contains either exactly one triangle (3-clique) or exactly zero triangles.
We generate it by sampling both random (nodes are connected to other random nodes) and $k$ nearest neighbor graphs (nodes are sampled as points on a 2D plane and then connected to their neighbors).
Then the edges which belong to triangles are removed from the graph until only one or zero triangles remain.
An example of data is shown in Figure~\ref{fig:triangles}.
The graphs themselves are rather complicated with usually multiple cycles of different lengths present in a graph.

\subsection{Clique Distance Dataset}

The second synthetic dataset checks whether an approach can find high-order patterns in the data.
We generate a base BA-graph \cite{Albert:2002:rmp} and then attach two cliques to two distinct nodes of the base graph.
Connection parameter $m$ is set to $k - 2$, where $k$ is the clique size.
This enforces a fact that the BA graphs can contain cliques only up to the $k - 1$ and there is no noise in the data.
We sample the number of nodes uniformly from 5 to 20 and set the clique size to 4.
An example of data is shown on Figure~\ref{fig:clique_dist}.
Graphs which has the shortest distance between the cliques below a threshold get a label of 0, otherwise 1.
We set the distance threshold to 4.


\section{Experiments} 
\label{sec:experiments}

We perform experiments on both synthetic and real-world data.
Real-world datasets are standard benchmarks for the graph classification tasks, and synthetic datasets highlight specific capabilities of the proposed model.
All experiments were performed with 10-fold cross-validation with a fixed random seed and we report average scores.

We use three types of graph neural networks as baselines: original \textbf{GCN} \cite{scarselli2008graph}, \textbf{GIN} \cite{Xu2019}, \textbf{GAT} \cite{Velickovic2018}.
We use the GNN implementations provided by BenchmarkingGnns project \cite{dwivedi2020benchmarkgnns}.
As a non-neural baseline we use SVM \cite{svm-paper} with WL kernel \cite{shervashidze2011weisfeiler}.
We label it as \textbf{SVM+WL}.
For synthetic datasets, we also try graphlet sampling kernel \cite{pmlr-v5-shervashidze09a}.
We label it as \textbf{SVM+Graphlets}.
We use GraKel \cite{JMLR:v21:18-370} framework for kernel implementations.
We use six layers for all GNNs and leave other parameters at 100k settings of the provided configuration files.

We use three types of input node features for GNNs.
The first one is uniform: each node is initialized with the same feature vector.
It is not denoted specifically.
The second type adds node degree features encoded as a single scalar.
We denote it as \textbf{+D}.
The last one also adds node ID features, denoted as \textbf{+ID} in one-hot encoding.
We do not use specific node features for SVM kernels.

\subsection{Node Index Shuffling + Transformer Approach} 
\label{sub:our_approach}

We also include a proposed method, which uses a Transformer \cite{NIPS2017_7181} model as a state update equation.
We use Transformer in two configurations.
The first variant uses Transformer as is, effectively treating graph as fully connected.
We denote it as \textbf{TF}.

The second variant, labeled as \textbf{TF+AM}, limits the self-attention to the graph adjacency matrix.
Limiting the Transformer's self-attention to a graph adjacency matrix makes the overall method a hybrid between the message-passing approach and embedding topology into the embeddings.
For TF, we use four Transformer layers with four heads and a hidden layer dimension of 32 for self-attention.
In the TF+AM setting, we add graph adjacency matrix masking to two first Transformer layers.
In the AM4 setting, we use adjacency matrix masking in all four Transformers, effectively ending up with a GNN-like network.

We use a concatenation of local node embeddings and the sum of adjacent node embeddings as initial features.
We also permute the numbering of individual nodes, while keeping the graph topology intact.
We hypothesize that the node shuffling makes it capture graph topology directly in the embeddings.
This method is inspired by a tensor decomposition approach and in the Appendix we provide an extension for handling node and edge labels as well.
We also report variants without index shuffling (\textbf{-IS}) and a fully-connected neural network baseline (\textbf{FCNN}).


\subsection{Synthetic Data}

\begin{wraptable}{r}{5cm}
\begin{tabular}{lrrrr}
Method & NCI1 & MUTAG \\
\toprule
GCN & 79.0 & 87.5 \\
GIN & 81.0 & 90.0 \\
GAT & 78.7 & 87.1 \\
SVM+WL & 78.4 & 82.8 \\
\midrule
TF & 75.9 & 86.5 \\
TF+AM & 82.6 & 88.7 \\
\end{tabular}
\caption{Test F1 score, \% on real-world datasets}
\label{tab:real_datasets}
\end{wraptable}

Firstly, we test all the algorithms on the proposed synthetic data.
Table~\ref{tab:synch_datasets} shows the prediction results of the tested algorithms.
In the Triangles dataset, GIN seems to capture some features even in the default setting, but both GCN and GAT completely fail to learn anything producing the same label for all graphs.
When adding node degree features, the accuracy of GCN and GAT significantly increases, but node IDs do not add significant further impact.
SVM with WL kernel also fails to achieve 100\% accuracy.
WL kernel focuses on labels and is weak for unlabeled tasks.
Graphlet kernel, on the other hand, achieves almost perfect accuracy.
We believe that its failures come from the stochastic nature of the graphlet sampling.
Both variants of the proposed method achieve perfect accuracy.

Graphlet kernel-based SVM in theory should be a perfect match for this problem because it can capture triangles directly.
Other methods have to perform cycle detection.
A method should develop a way of tracking node identities to be able to detect cycles.
However, GNNs, even if provided with node identity information, can not detect triangles perfectly.
On the other hand, our proposed method can.
We hypothesize that label shuffling can be a strong data augmentation for this task, however, as an additional experiment, even without label shuffling the TF+AM setting can achieve almost perfect accuracy ($>$99\%).
Additionally, it seems that attention-based methods (TF variants and GAT) perform better in detecting topology structures.

The Clique distance dataset has the picture for GNNs similar to the Triangles problem with the default setting, but GIN and GAT can solve the problem with additional features while GCN still struggles with it.
Node identities do not seem to be of much help for it as well.
SVMs perform poorly on this task because their kernels do not have enough representative power.
Our method in its basic form struggles with this dataset.
Occasionally it can learn good topological embeddings and perform well, but the situation is not stable.
However, adding an adjacency mask even to one Transformer layer changes the situation drastically, and the experimental setting of two adjacency-restricted layers and two full Transformer layers is always able to solve the task.

\subsection{Real-World Data}

We additionally check whether the proposed method keeps its performance with real-world datasets as well.
As seen in Table~\ref{tab:real_datasets}, we select standard benchmark graph datasets: NCI1, MUTAG \cite{shervashidze2011weisfeiler}.
We use the parameters for GNNs provided by \cite{dwivedi2020benchmarkgnns}.
We also employ loss weighting for unbalanced datasets for all methods.
We do not use Graphlet kernel SVM for real-world data.
The general picture is the same for all real-world datasets: GCN and plain TF being the lowest, with GIN, GAT, and TF+AM being higher.


\section{Conclusion} 
\label{sec:conclusion}

We propose two seemingly simple synthetic datasets which show that GNNs can not extract graph topology features reliably.
We also propose a method that reliably extracts graph topology features, at least from proposed datasets.
The future work would be to investigate how it would be possible to modify existing GNN methods to extract topology features or find new regimes where GNNs break.

\section{Acknowledgments}

We thank Shotaro Yano for his help with generating the datasets.
We also thank Masafumi Shingu for discussions on our approach for handling topological problems.

\bibliography{nndt_gtrl.bib}
\bibliographystyle{iclr2021_workshop}

\appendix

\section{Creating Initial Topological Embeddings} 
\label{sec:propsed_approach}

The goal of our approach is to create initial embeddings for the input data.
It uses the intuition that the tensor decomposition approaches implicitly capture topology information.

In this section, we start by describing the overall flow of the complete neural network.
Then we briefly introduce preliminaries on tensor decompositions and operations on sparse matrices and tensors followed by the detailed description of our method.

\subsection{Outline}

\begin{figure}[tb]
	\centering
	\includegraphics[width=0.3\columnwidth]{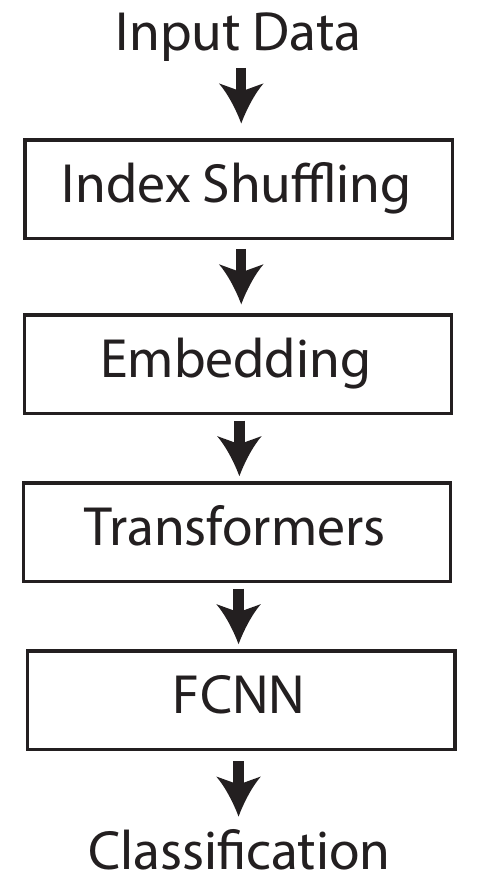}
	\caption{Graph classification outline}
	\label{fig:outline}
\end{figure}

\begin{figure}[tb]
	\centering
	\includegraphics[width=0.15\columnwidth]{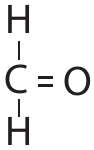}
	\caption{Formaldehyde chemical structure}
	\label{fig:formaldehyde}
\end{figure}


\begin{figure}[tb]
\centering
\begin{tabular}{lllll|l}
	0 & 1 & C & H & 0 & 1 \\
	0 & 2 & C & H & 0 & 1 \\
	0 & 3 & C & O & 1 & 1 \\
	1 & 0 & H & C & 0 & 1 \\
	2 & 0 & H & C & 0 & 1 \\
	3 & 0 & O & C & 1 & 1 \\
\end{tabular}
\caption{
Formaldehyde sparse tensor representation.
Table columns represent tensor modes.
Modes 1 and 2 are topology.
Modes 3 and 4 are node labels (atom types) for modes 1 and 2 respectively.
Mode 5 is the edge label (atomic bond type).
The last column is the entry weight.
}
\label{fig:tensor_form}
\end{figure}

The whole process of graph classification is shown in Figure~\ref{fig:outline}.
The process of node classification follows the same flow.

For our approach, the data is assumed to be in the tensor form.
For the graph and relational data, the input tensor $X$ is usually very sparse and the tensor is stored in a sparse form.
Figure~\ref{fig:tensor_form} shows the sparse tensor form of the formaldehyde (H$_2$OC) molecule, which is shown on Figure~\ref{fig:formaldehyde} as the compound (graph) form.
For clarity, we did not replace atom types with indices.
The tensor data contains label modes (columns) and topology modes.
Their precise definition is given later.

The first step of our processing is to shuffle the indices of the topology modes.
That forces the neural network to learn permutation invariant weight matrices for the topology modes.

The second step creates the initial embeddings for the data.
It consists of a sequence of mode projection and mode embedding operations.
These operations are described later.

The next step is several Transformer layers.
Their role is similar to message passing in graphs, however, passing directions are learned.

Finally, a small fully-connected neural network is used with the resulting representations from the Transformers.

{ 
\newcommand{\setR}{\mathbb{R}}
\newcommand{\nnz}{\operatorname{nnz}}

\subsection{Tensor Decomposition and Mode Products}

A tensor decomposition of a order-$n$ tensor $X \in \setR^{I_1 \times I_2 \times \cdots \times I_n}$ results in a core tensor $\bar{X} \in \setR^{J_1 \times J_2 \times \cdots \times J_n}$ and a set of matrices $W_i \in \setR^{I_i \times J_i}, i = 1...n$.
Individual tensor directions are referred to as \emph{modes}.
The core tensor could be computed by a sequence of \emph{mode products} between the tensor and the matrices.

A mode product for the mode $k$ with the matrix $W_k \in \setR^{I_k \times J_k}$ is defined as
\begin{equation*}
	(X \times_k W_k)_{i_1 i_2 ... i_{k-1} j_k i_{k+1} ... i_n} =
	\sum_{i_k=1}^{I_k} x_{i_1 i_2 ... i_n} w_{i_k j_k},
\end{equation*}
where $x_{i_1...i_n}$ is an element of the tensor $X$ and $w_{i_k j_k}$ is an element of the matrix $W_k$.
The resulting tensor $(X \times_k W_k)$ has dimension
\begin{equation*}
	(X \times_k W_k) \in \setR^{I_1 \times \cdots \times I_{k-1} \times J_k \times I_{k+1} \times \cdots \times  I_n}.
\end{equation*}
Refer to \cite{Kolda2009} for a more detailed explanation.
From the mathematical point of view, the order of mode products can be arbitrary, but the actual implementation has to define a concrete sequence of the products.
In the following discussion, we fix the computation order from the last tensor mode $n$-th to the first one.

\subsection{Sparse Matrices}

There exist many sparse matrix formats focusing on different aspects of computational efficiency, but most of which are derived from the coordinate sparse matrix format.
A matrix in the coordinate sparse format is represented as a two-element tuple $(I, V)$ where $I$ is $2 \times n$ array of indices and $V$ is a vector of length $n$ which contains the values of the matrix.
$I$ and $V$ have their rows in correspondence, meaning that $I_i$ is the coordinate of the $V_i$.
Matrix elements for which the indices are not contained in $I$ are assumed to be zero.

\begin{figure}[tb]
	\centering
	\includegraphics[width=0.95\columnwidth]{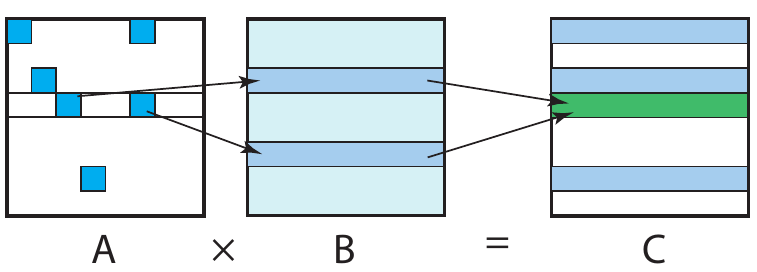}
	\caption{
	Sparse-dense matrix multiplication.
	Matrix A is sparse, matrix B is dense.
	For each non-zero row of the sparse matrix, the row elements are multiplied with the respective row of the dense matrix, forming the rows of the result matrix.
	A dense subelement of $A$ is a scalar, but a dense subelement of $C$ is a vector.
	}
	\label{fig:sd_matmul}
\end{figure}

A result of matrix multiplication of a sparse matrix and a dense matrix is denser than the original sparse matrix.
Sparse-dense matrix multiplication is illustrated in Figure~\ref{fig:sd_matmul}.

\subsection{Mixed Tensor Representation}

For our needs, we extend the coordinate matrix representation to handle tensors as well.
Note that the resulting matrix of the sparse-dense matrix multiplication, as described Figure~\ref{fig:sd_matmul}, has vectors (order-1 tensors) instead of scalars (order-0 tensors) as \emph{dense sub-elements}.
Computing the mode product for a sparse tensor, being a matrix multiplication in nature, also has the effect of increasing the order of the dense subelements.
The mixed tensor representation helps us to capture this fact.

A simple coordinate sparse tensor representation consists of an index table and the corresponding non-zero subtensor elements.
It is a straightforward extension of the coordinate sparse matrix format.
A tensor $X$ in a sparse form is a tuple $(I, V)$.
Rows of the index table $I$ are the coordinates that correspond to the vector elements $V$.
Columns of the index tables are the tensor modes.
We refer to tensor modes as the numbers in parenthesis, e.g., the fifth mode would be referred to as $(5)$.

When we use tensors instead of scalars as the elements of $V$ in a sparse representation, we end up with a mixed tensor representation.
A row $I_i = [I_i^{(1)}, \dots, I_i^{(n)}]$ of the index table $I$ points to a dense subtensor unit $V_i$ of the whole tensor.
Similar to other sparse formats, zero elements are not stored.
Subtensors, however, can contain accidental zeros.

\subsection{Mode Product in the Mixed Representation}

A mode product of a tensor $X$ in the mixed representation with the matrix $W$ can be formulated as index partitioning, tensor outer product, finally followed by a summation.
It is a natural extension of the sparse-dense matrix multiplication.
The difference is that the product of a dense matrix row with the sparse matrix element is replaced by an outer product for the mixed tensor representation case.
Without restricting generality, suppose we want to compute a mode product with the last ($n$-th) mode.
First, we partition index rows as
\begin{equation}
	[I_i^{(1)}, \dots, I_i^{(n-1)} | I_i^{(n)}] = [\hat{I}_i | I_i^{(n)}].
\end{equation}
The output index table $I'$ will be formed from the values $\hat{I}_i$.
Because $\hat{I}_i$ contains one tensor mode less than $I_i$, the rows $\hat{I}_i$ are not unique in general.
The unique rows $\hat{I}_i$ will form the index table $I'$ of the mode product result tensor $X'$.
The respective values of $V'$ can be computed as a summation over the outer product of rows of $V_i$ with the $I^{(n)}_i$-th rows of the weight matrix $W$:
\begin{equation}
	V'_j = \sum_{i \in \nnz(I'_j)} V_i \otimes W_{I^{(n)}_i}.
	\label{eq:mode_product_mixed}
\end{equation}
This is an equation of the mode product of a tensor $X$ in the mixed representation with a matrix $W$ over the last sparse mode.
The summation is done over a set of indices where the partitioned elements $\hat{I}_i$ are the same and we denote them as
\begin{equation}
	\nnz(I'_j) = \{i: \hat{I}_i = I'_j\}.
	\label{eq:nnz_j}
\end{equation}
The resulting index table $\hat{I}$ becomes one mode smaller and subtensors $\hat{V}_j$ become one mode larger.
In the mixed representation, the mode product converts a single tensor mode from sparse to dense.
The whole computation of a core tensor could be thought of as a sequence of these operations, starting with fully sparse representation and ending with a single dense tensor.

\subsection{Handling Labels}

Tensors arising from graphs often contain labels that result in label modes.
A \emph{label mode} contains a single non-zero value when other modes are fixed, or using the definitions from eq.~\ref{eq:mode_product_mixed}, $\forall j: |\nnz(I'_j)| = 1$.
The summation in eq.~\ref{eq:mode_product_mixed} goes away and the operation becomes identical to an outer product for each subtensor value.
This outer product can be thought of as adding new information from the label embedding vector ($W_{I^{(n)}_i}$) to the representation $V$.
The outer product with a vector increases the subtensor order, exponentially increasing the memory size of subtensors.

To make the computations efficient, when handling labels, we would like to increase the memory usage linearly while keeping the same information like the mode product.
To do that, we first change the mixed representation to use \emph{sets} of subtensor values $V = \{V^1, ..., V^k\}$ instead of simple tensors.
The subtensors can also be of a different order.

For handling labels, we modify the current value set $V$ without changing the indices:
\begin{equation}
	\hat{V} = V \cup \left\{ [W_{I^{(n)}_i}] \right\}.
\end{equation}
Its intuition is to \emph{append} the embedding vector to the current representation.
This new element consists of the rows of $W$ indexed by the label mode values $I^{(n)}_i$.

The final representation is going to be an input of the neural network, so we flatten and then concatenate the elements of the set $V$ creating the neural network inputs.
In this sense, the outer product (of the mode product) and the vector concatenation should contain the same information, but using a different number of values.
Remember that the outer product of two vectors is a rank-1 matrix.

\subsection{Initial Embeddings for Data}

For the graph data, we would like to use the proposed scheme for creating initial node representations, which then would be fed to the downstream neural network.
In general, we assume that the data contains at least two non-label (topology) modes and several label modes (see Figure~\ref{fig:formaldehyde}).
Graph data would contain exactly two topology modes which correspond to the graph adjacency matrix.
We also assume that label modes depend on one or more topology modes: label mode values can be uniquely determined by a subset of other modes.
For example, in graph data node labels will depend on a single node topology mode and edge labels would depend on both topology modes.
We formulate the process of computing initial embeddings as accepting the input data in the sparse tensor format and producing the vector representations for the \emph{main mode}.
For the graph data, we assume that one of the topology modes is the main mode.

To recap: we defined two operations (mode product and label embedding), each operation takes a tensor in mixed representation with $n$ sparse modes and $k$ dense modes and output a tensor in mixed representation with $n - 1$ sparse modes and $k + 1$ dense modes.
Each operation has a parameter matrix that is learned.
We create the initial node embeddings with the following algorithm.
\begin{enumerate}
	\item Chose the main topology mode
	\item Reorder modes: main topology modes, other topology modes, labels
	\item Topology modes are processed by mode products, label modes are processed by mode embeddings
	\item Label modes should be processed as late as possible
	\item Label modes must be processed before the dependent topology modes
	\item Process the main topology mode with a mode embedding that has its parameters shared with the mode product for the other mode product
\end{enumerate}

\end{document}